%% file: emnlp2018.tex
\documentclass[11pt,a4paper]{article}
\usepackage[hyperref]{emnlp2018}
\usepackage{times}
\usepackage{latexsym}

\usepackage{graphicx}
\usepackage{url}
\usepackage{adjustbox}
\graphicspath{ {images/} }
\usepackage[font=small,labelfont=bf]{caption}
\usepackage[utf8]{inputenc}
\usepackage{pgfplots}
\usepackage{todonotes}
\usepackage{bm}
\usepackage{amsbsy}
\usepackage{ulem}
\usepackage{pifont}
\usepackage[symbol]{footmisc}

\usepackage[T1]{fontenc}
\usepackage[font=small,labelfont=bf]{caption}
\usepackage{url}
\usepackage[lofdepth,lotdepth]{subfig}
\usepackage{tabularx}
\usepackage{tablefootnote}
\DeclareCaptionLabelFormat{andtable}{#1~#2  \&  \tablename~\thetable}

\newcommand\confname{EMNLP 2018}

\newcommand{\cm}[1]{\textcolor{red}{#1}}
\newcommand{\rep}[1]{\textcolor{blue}{#1}}

\title{Instructions for \confname{} Proceedings}

\author{First Author \\
  Affiliation / Address line 1 \\
  Affiliation / Address line 2 \\
  Affiliation / Address line 3 \\
  {\tt email@domain} \\\And
  Second Author \\
  Affiliation / Address line 1 \\
  Affiliation / Address line 2 \\
  Affiliation / Address line 3 \\
  {\tt email@domain} \\}

\date{}
\title{emrQA: A Large Corpus for Question Answering on \\Electronic Medical Records}

\author{Anusri Pampari\Thanks{Part of this work was done during an internship at IBM }  \ \ \    Preethi Raghavan$^\dagger$\textsuperscript{\ding{168}}     \ \ \  Jennifer Liang$^\dagger$ and Jian Peng$^*$$^\ddagger$\\
           \textsuperscript{\ding{168}}MIT-IBM Watson AI Lab, Cambridge, MA \\
         $^\dagger$IBM TJ Watson Research Center, Yorktown Heights, NY\\
$^*$Dept. of Computer Science, University of Illinois Urbana Champaign, IL\\
$^\ddagger$Carle Illinois College of Medicine, University of Illinois Urbana Champaign, IL\\
            {\tt $^*$\{pampari2,jianpeng\}@illinois.edu $^\dagger$\{praghav,jjliang\}@us.ibm.com}\\
}
\date{}
\aclfinalcopy
\begin{document}
\maketitle

\input{abstract}
\input{introduction}
\input{related_work}
\input{generation_framework}
\input{emrqa_generation}
\input{emrqa_analysis}
\input{emrqa_baselines}
\input{discussion}
\input{conclusion}

\vspace{-0.05in}
\section*{Acknowledgments}
\vspace{-0.04in}
This project is partially funded  by  Sloan  Research  Fellowship, PhRMA  Foundation  Award  in  Informatics,  and  NSF  Career  Award  (1652815). The authors would like to thank Siddharth Patwardhan for his valuable feedback in formatting the paper.

\bibliography{emnlp2018}
\bibliographystyle{acl_natbib_nourl}

\end{document}

%% file: abstract.tex
\begin{abstract}
%Building a question answering (QA) system for answering physician questions on a patient's electronic medical record (EMR) requires vast amounts of expertly annotated data that is extremely difficult to obtain
We propose a novel methodology to generate domain-specific large-scale question answering (QA) datasets by re-purposing existing annotations for other NLP tasks. We demonstrate an instance of this methodology in generating a large-scale QA dataset for electronic medical records by leveraging existing expert annotations on clinical notes for various NLP tasks from the community shared i2b2 datasets\footnote[4]{\label{note1}\href{https://www.i2b2.org/NLP/DataSets/}{https://www.i2b2.org/NLP/DataSets/}}. The resulting corpus (emrQA) has 1 million question-logical form and 400,000+ question-answer evidence pairs. We characterize the dataset and explore its learning potential by training baseline models for question to logical form and question to answer mapping.
\end{abstract}
\vspace{-0.4cm}

%% file: introduction.tex
\section{Introduction}
%\vspace{-0.06in}
%\todo{you comment like this}
%\cm{you can suggest removing words like this}
%\rep{you can suggest addition edits like this}
Automatic question answering (QA) has made big strides with several open-domain and machine comprehension systems built using large-scale annotated datasets \cite{voorhees1999trec, ferrucci2010building, rajpurkar2016squad, joshi2017triviaqa}. However, in the clinical domain this problem remains relatively unexplored. Physicians frequently seek answers to questions from unstructured electronic medical records (EMRs) to support clinical decision-making \cite{demner09}.
%EMRs are longitudinal records of patient health information captured in unstructured clinical notes (progress notes, discharge summaries etc.) and structured vocabularies \cite{demner09}. 
But in a significant majority of cases, they are unable to unearth the information they want from EMRs \cite{tang1994traditional}. Moreover to date, there is no general system for answering natural language questions asked by physicians on a patient's EMR (Figure \ref{note}) due to lack of large-scale datasets \cite{ragpatamia}.
\begin{table}[t]
\footnotesize
\renewcommand{\arraystretch}{1.2}
  \begin{tabular}{p{7.5cm}}
  \hline
Record Date: 08/09/98\\
\hline
\rep{08/31/96 ascending aortic root replacement with homograft with omentopexy.} The patient continued to be hemodynamically stable making good progress. Physical examination: \cm{BMI: 33.4  Obese, high risk}. Pulse: 60. resp. rate: 18 \\
\\
\textbf{Question:} Has the patient ever had an abnormal BMI?\\
\textbf{Answer:} \cm{BMI: 33.4  Obese, high risk} \\
\textbf{Question:} When did the patient last receive a homograft replacement ? \\
\textbf{Answer:} \rep{08/31/96 ascending aortic root replacement with homograft with omentopexy.} \\
 \hline
  \end{tabular}
\captionof{figure}{Question-Answer pairs from emrQA clinical note.}
  \label{note}
 \vspace{-0.3in}
\end{table}
EMRs are a longitudinal record of a patient's health information in the form of unstructured clinical notes (progress notes, discharge summaries etc.) and structured vocabularies. Physicians wish to answer questions about medical entities and relations from the EMR, requiring a deeper understanding of  clinical notes. While this may be likened to machine comprehension, the longitudinal nature of clinical discourse, little to no redundancy in facts, abundant use of domain-specific terminology, temporal narratives with multiple related diseases, symptoms, medications that go back and forth in time, and misspellings, make it complex and difficult to apply existing NLP tools \cite{demner09,ragpatamia}. Moreover, answers may be implicit or explicit and may require domain-knowledge and reasoning across clinical notes. Thus, building a credible QA system for patient-specific EMR QA requires large-scale question and answer annotations that sufficiently capture the challenging nature of clinical narratives in the EMR.  However, serious privacy concerns about sharing personal health information \cite{devereaux2013use,krumholz2016data}, 
and the tedious nature of assimilating answer annotations from across longitudinal clinical notes, makes this task impractical and possibly erroneous to do manually \cite{lee2017big}. 

%Thus, we define the task where a physician seeks answers to questions about the patient's health from the longitudinal EMR ``patient-specific EMR QA",  \cite{ragpatamia}. Thus, building a credible QA system for patient-specific EMR QA requires large-scale question and answer annotations that sufficiently capture the challenging nature of clinical narratives in the EMR.   

 \begin{figure*}[t]
    \centering
    \includegraphics[width=\textwidth, height=2in]{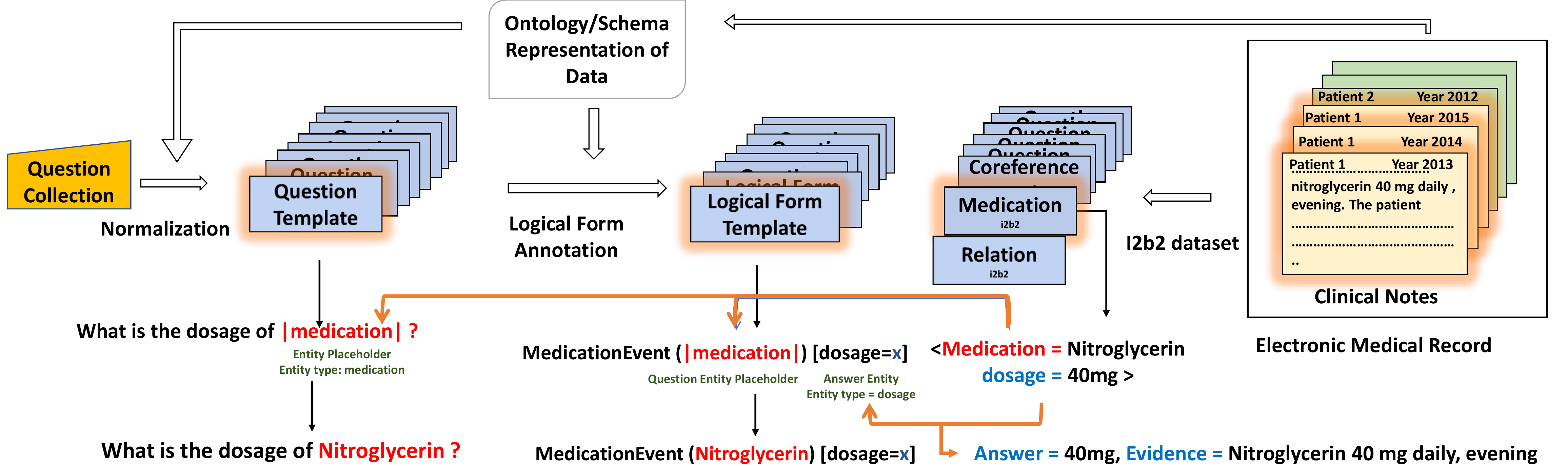}
    \caption{Our QA dataset generation framework using existing i2b2 annotations on a given patient's record to generate a question, its logical form and answer evidence. The highlights in the figure show the annotations being used for this example. }
    \label{fig:mesh1}
\vspace{-0.15in}
\end{figure*}

In this work, we address the lack of any publicly available EMR QA corpus by creating a large-scale dataset, emrQA, using a novel generation framework that allows for minimal expert involvement and re-purposes existing annotations available for other clinical NLP tasks (i2b2 challenge datasets \cite{guo2006identifying}). 
%The resource re-purposed is the community shared i2b2 datasets \cite{guo2006identifying} which comprise of expert annotations for various complex clinical NLP tasks (e.g., co-reference resolution, temporal relation, concept and relation extraction).  
%The proposed framework, uses the annotations from the i2b2 datasets to populate expert generated question templates and logical forms.
%The annotations serve as a proxy-expert in populating question templates and generating, answers and logical forms. 
The annotations serve as a proxy-expert in generating questions, answers, and logical forms. Logical forms provide a human-comprehensible symbolic representation, linking questions to answers, and help build interpretable models, critical to the medical domain \cite{ davis1977production, vellido2012making}. We analyze the emrQA dataset in terms of question complexity, relations, and the reasoning required to answer questions, and provide neural and heuristic baselines for learning to predict question-logical forms and question-answers.
%We also implement heuristic and neural baselines for both learning to map questions to logical forms as well as questions to answers and demonstrate that emrQA dataset requires models for deeper question understanding and answer retrieval.

The main contributions of this work are as follows:
\begin{itemize}
\item A novel framework for  systematic generation of domain-specific large-scale QA datasets that can be used in any domain where manual annotations are challenging to obtain but limited annotations may be available for other NLP tasks.
\item The first accessible patient-specific EMR QA dataset, emrQA\footnote{\href{https://github.com/panushri25/emrQA}{https://github.com/panushri25/emrQA}, scripts to generate emrQA from i2b2 data. i2b2 data is accessible by everyone subject to a license agreement.}, consisting of 400,000 question-answer pairs and 1 million question-logical form pairs. The logical forms will allow users to  train and benchmark interpretable models that justify answers with corresponding logical forms. 
\item Two new reasoning challenges, namely arithmetic and temporal reasoning, that are absent in open-domain datasets like SQuAD \cite{rajpurkar2016squad}.
\end{itemize}
%\vspace{-0.1in}

%% file: related_work.tex
\section{Related Work}
\vspace{-0.04in}
Question Answering (QA) datasets are classified into two main categories: (1) machine comprehension (MC) using unstructured documents, and (2) QA using Knowledge Bases (KBs).
%Question Answering (QA) has been a longstanding goal of NLP with datasets created over the years in two main categories. 1) Machine Comprehension (MC) using unstructured documents and 2) QA using Knowledge Bases (KBs).
% Give a context of semantic frames 	\cite{johnson2000framenet,niwattanakul2013using,zajac2001towards} Machine comprehension datasets like  SQuAD \cite{rajpurkar2016squad} and NewsQA \cite{trischler2016newsqa} have 100K  natural language questions with answers collected from short passages using crowd-sourcing.   TriviaQA \cite{joshi2017triviaqa} with 650K trivia  enthusiast authored question-answer pairs and QUASAR-T with 430K \cite{dhingra2017quasar} trivia questions and their answers obtained from various internet sources have large text documents  (for e.g., Wikipedia.) drawn from the search results as they rely on the fact being redundant across the.

MC systems aim to answer any question that could be posed against a reference text. %The paradigm of exploiting text for answering questions started in the early 1990s \cite{kupiec1993murax}.
%With the advent of web, access to text resources became abundant and cheap. 
Recent advances in crowd-sourcing and search engines have resulted in an explosion of large-scale (100K) MC datasets for factoid QA, having ample redundant evidence in text \cite{rajpurkar2016squad, trischler2016newsqa, joshi2017triviaqa, dhingra2017quasar}. On the other hand, complex domain-specific MC datasets such as MCTest \cite{richardson2013mctest}, biological process modeling \cite{berant2014modeling}, BioASQ \cite{tsatsaronis2015overview}, InsuranceQA \cite{feng2015applying}, etc  have been limited in scale (500-10K) because of the complexity of the task or the need for expert annotations that cannot be crowd-sourced or gathered from the web. In contrast to the open-domain, EMR data cannot be released publicly due to privacy concerns \cite{vsuster2017short}. Also, annotating unstructured EMRs requires a medical expert who can understand and interpret clinical text. Thus, very few datasets like i2b2, MIMIC \cite{johnson2016mimic} (developed over several years in collaboration with large medical groups and hospitals), share small-scale annotated clinical notes.  In this work, we take advantage of the limited expertly annotated resources to generate emrQA. 

%The objective of our work is to take advantage of the limited expertly annotated resources to generate emrQA.
%; our framework is capable of generating a large scale corpus with reduced burden on experts.cite{ %The LFs produce answers when executed on any structured schema  %or a schematic \todo{check if correct}  representation of free text. 
%\cite{auer2007dbpedia,yang2015wikiqa,berant2014modeling,johnson2000framenet}. Traditional semantic parsers require annotated LFs as supervision 
KB-based QA datasets, used for semantic parsing, are traditionally limited by the requirement of annotated question and logical form (LF) pairs for supervision where the LF are used to retrieve answers from a schema \cite{ cai2013large, lopez2013evaluating, bordes2015large}.  \newcite{roberts2016annotating} generated a corpus by manually annotating LFs on 468 EMR questions (not released publicly), thus limiting its ability to create large scale datasets. In contrast, we only collect LFs for question templates from a domain-expert - the rest of our corpus is automatically generated.
%and re-purposes existing annotations for other NLP tasks as a proxy expert.

Recent advances in QA combine logic-based and neural MC approaches to build hybrid models \cite{usbeck2015hawk, feng2016hybrid, palangi2018question}. These models are driven to combine the accuracy of neural approaches \cite{hermann2015teaching} and the interpretability of the symbolic representations in logic-based methods \cite{gaomachine, chabierski2017logic}. Building interpretable yet accurate models is extremely important in the medical domain \cite{shickel2017deep}. We generate large-scale ground truth annotations (questions, logical forms, and answers) that can provide supervision to learn such hybrid models. 
Our approach to generating emrQA is in the same spirit as \newcite{su2016generating}, who generate graph queries (logical forms) from a structured KB and use them to collect answers. In contrast, our framework can be applied to generate QA dataset in any domain with minimal expert input using annotations from other NLP tasks. %They then employ human annotators to convert graph queries into questions. 
%\cm{In contrast, our framework, with minimal expert input, can be applied to generate any new QA dataset, provided annotations for certain fundamental NLP tasks are at our disposal.}
%to query both unstructured and structured text provided NLP annotations for certain other tasks are at our disposal.
%\todo{JL: this last sentence is a little confusing - suggest rewording} %any data whether both structured or unstructured, domain-specific or open domain and can leverage existing expertise and annotations to generate a QA corpus. 

\begin{table}[t]
\footnotesize
\centering
  \begin{tabular}{@{\hspace{-0.1\tabcolsep}} l  @{\hspace{0.1\tabcolsep}} c  @{\hspace{0.5\tabcolsep}} c  @{\hspace{0.5\tabcolsep}} c @{\hspace{0.1\tabcolsep}} | p{2.5cm} @{\hspace{-2.8\tabcolsep}} c @{\hspace{-0.5\tabcolsep}}}
  \hline 
  \textbf{Datasets} & \textbf{\#QA} & \textbf{\#QL}  & \textbf{\#notes} & \textbf{Property} & \textbf{Stats.}\\
\hline
 Relations &  141,243 & 1,061,710 & 425 &  Question len. & 8.6\\ 
Medications & 255,908 & 198,739 & 261 &Evidence len.& 18.7\\ 
Heart disease & 30,731 & 36,746 & 119 & LF len. & 33 \\ 
Obesity & 23,437 & 280 &  1,118&  Note len.& 3825\\ 
Smoking & 4,518 & 6 & 502 &  \# of evidence& 1.5\\
emrQA & 455,837 & 1,295,814 & 2,425 & \# Ques. in note& 187\\ 
\hline
  \end{tabular}
  \caption{(left) i2b2 dataset distribution in emrQA, and (right) emrQA properties with length in tokens, averaged}
  \label{tab:3}
 \vspace{-0.2 in}
\end{table}
%\vspace{-0.11in}

%% file: generation_framework.tex
\vspace{-0.05in}
\section{QA Dataset Generation Framework}
\label{frames}
\vspace{-0.04in}
Our general framework for generating a large-scale QA corpus given certain resources consists of three steps: (1) collecting questions to capture domain-specific user needs, followed by normalizing the collected questions to templates by replacing entities (that may be related via binary or composite relations) in the question with placeholders. The entity types replaced in the question are grounded in an ontology like WordNet \cite{miller1995wordnet}, UMLS \cite{bodenreider2004unified}, or a user-generated schema that defines and relates different entity types. (2) We associate question templates with expert-annotated logical form templates; logical forms are symbolic representations using relations from the ontology/schema to express the relations in the question, and  associate the question entity type with an answer entity type. 
%Hence, obtaining expert manual annotations in complex domains  is infeasible as it is tedious to expert-annotate answers that may be found across long document collections (e.g., longitudinal EMR)
(3) We then proceed to the important step of re-purposing existing NLP annotations to populate question-logical form templates and generate answers. QA is a complex task that requires addressing several fundamental NLP problems before accurately answering a question. Hence, obtaining expert manual annotations in complex domains  is infeasible as it is tedious to expert-annotate answers that may be found across long document collections (e.g., longitudinal EMR) \cite{lee2017big}. Thus, we reverse engineer the process where we reuse expert annotations available in NLP tasks such as entity recognition, coreference, and relation learning, based on the information captured in the logical forms to populate entity placeholders in templates and generate answers. Reverse engineering serves as a proxy expert  ensuring that the generated QA annotations are credible. The only manual effort is in annotating logical forms, thus significantly reducing expert labor. Moreover, in domain specific instances such as  EMRs, manually annotated logical forms allow the experts to express information essential for natural language understanding such as domain knowledge, temporal relations, and negation \cite{gaomachine,chabierski2017logic}. This knowledge, once captured, can be used to generate QA pairs on new documents, making the framework scalable.

\vspace{-0.05in}

%% file: emrqa_generation.tex
\section{Generating the emrQA Dataset}
\vspace{-0.04in}
%\vspace{-0.08in}
\label{emrqa}
We apply the proposed framework to generate the emrQA corpus consisting of questions posed by physicians against longitudinal EMRs of a patient, using annotations provided by i2b2 (Figure \ref{fig:mesh1}). 
%\textbf{Question and Template Generation.}
%the application of the framework to generate emrQA corpus and also discuss its possible extensions and 
%\todo{discussing limitations of your approach, while being noble, might not be a good idea.}limitations.

 \begin{table}[t]
  %\hspace{0.50cm}
\footnotesize
  \begin{tabular}{l}
  \hline
%   \textbf{Question Paraphrases} \\
%   \hline
 How was the $|problem|$ managed ? \\
 How was the patient's $|problem|$ treated ? \\
 What was done to correct the patient's $|problem|$ ? \\
Has the patient ever been treated for a $|problem|$ ? \\
What treatment has the patient had for his $|problem|$ ?	 \\
Has the patient ever received treatment for $|problem|$ ? \\
What treatments for $|problem|$ has this patient tried ?\\
% What interventions were done for the patient's $|problem|$  \\
%   What methods has the patient tried to manage the $|problem|$  \\
%   What kind of treatments have been tried for the $|problem|$	 \\
\hline
  \end{tabular}
\caption{Paraphrase templates of a question type in emrQA.}
  \label{para-table}
\vspace{-0.1in} 
\end{table}

%\vspace{-0.1in} 
\subsection{Question Collection and Normalization}
%\vspace{-0.06in}
\label{generation-question}
%Avg. of dataset statistics
% \begin{table}[t]
% \centering
% \footnotesize
% \renewcommand{\arraystretch}{1.1}
%   \begin{tabular}{c c c}
%   \hline
% \textbf{Events} & \textbf{Argument} &\textbf{Attributes}   \\
\label{qtemplate}
We collect questions for EMR QA by, 1) polling physicians at the Veterans Administration for what they frequently want to know from the EMR (976 questions), 2) using an existing source of 5,696 questions generated by a team of medical experts from 71 patient records \cite{preethi} and %,  which is representative of common question topics  \cm{cite}. 
3) using 15 prototypical questions from an observational study done by physicians \cite{tang1994traditional}. 
%The objective is to capture the natural distribution of such a dataset by collecting questions from different expert-sources.
To obtain templates, the questions were automatically normalized by identifying medical entities (using MetaMap \cite{aronson2001effective}) in questions and replacing them with generic placeholders. The resulting $\sim$2K noisy templates were expert reviewed and corrected (to account for any entity recognition errors by MetaMap). We align our entity types to those defined in the i2b2 concept extraction tasks \cite{uzuner2010extracting, uzuner20112010} - \textit{problem, test, treatment, mode and medication}. E.g.,  The question \textit{What is the dosage of insulin?} from the collection gets converted to the template \textit{What is the dosage of $|medication|$?} as shown in Fig.\ref{fig:mesh1}. This process resulted in 680 question templates. We do not correct for the usage/spelling errors in these templates, such as usage of "pt" for "patient", or make the templates gender neutral in order to provide a true representation of physicians' questions.
%The frequency of occurrence of our templates in the question collection indicates that certain information needs are frequent followed long tail of less frequent ones (mean 4, std 21.14).
Further, analyzing these templates shows that physicians most frequently ask about test results (11\%), medications for problem (9\%), and problem existence (8\%). The long tail following this includes questions about medication dosage, response to treatment, medication duration, prescription date, etiology, etc. Temporal constraints were frequently imposed on questions related to tests, problem diagnosis and medication start/stop. 

\begin{figure}
\includegraphics[width=0.5\textwidth]{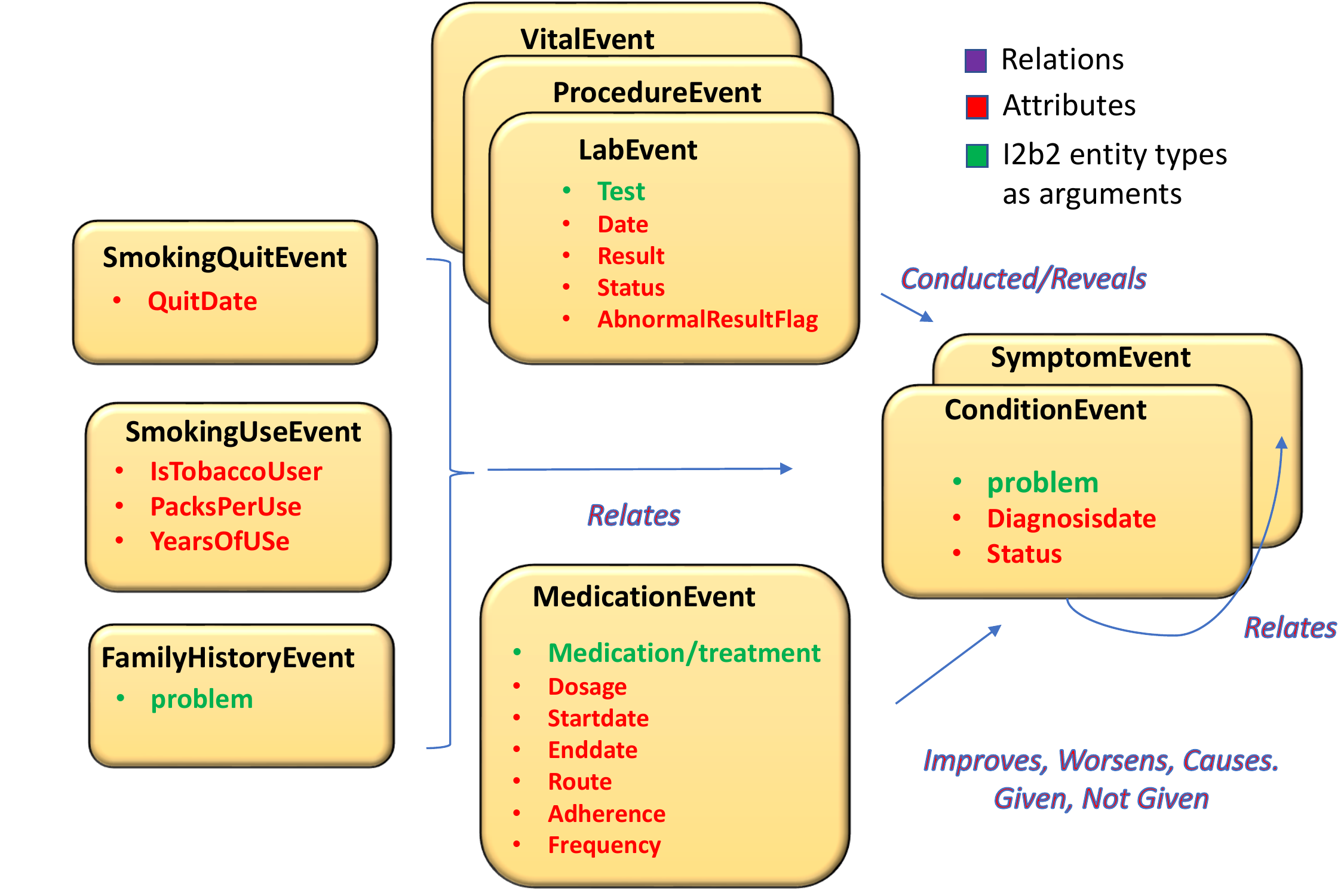}
\caption{Events, attributes \& relations in emrQA's logical forms. Events \& attributes accept i2b2 entities as arguments.}
\label{sem}
\vspace{-0.5cm}
\end{figure}
%\vspace{-0.3cm}
\subsection{Associating Templates w/ Logical Forms}
\label{sem-frames}
%\vspace{-0.06in}
%\rep{The relations expressed in the question were mapped to one of the relations defined as part of the i2b2 relations challenge shown in \cite{uzuner20112010}. These relations linking the entity types problem, treatment and test include \textit{treatment} \{improves, worsens, causes, is (not) administered for\} \textit{problem} and \textit{test} \{reveals, indicates\} \textit{problem}}. \todo{Marking this blue, since this part is not very clear, will need to look at to write it better}
%We now map the generated question templates to logical form templates. 

The 680 question templates were annotated by a physician with their corresponding logical form (LF) templates, which resulted in 94 unique LF templates. More than one question template that map to the same LF are considered paraphrases of each other and correspond to a particular question type (Table \ref{para-table}).
%The entities in the schema are collapsed to one of the i2b2 entity types i.e. diseases, symptoms correspond to problems, medications, procedures correspond to treatments and labs correspond to tests. \todo{can you make sure this is correct.}} \todo{marking this because very confusing-need to discuss
Logical forms are defined based on an ontology schema designed by medical experts (Figure \ref{sem}). This schema captures entities in unstructured clinical notes through medical events and  their attributes, interconnected through relations. We align the entity and relation types of i2b2 to this schema.   
%This schema models the entity-relation interactions in unstructured clinical notes as a graphical representation of medical events interconnected through relations. The events have attributes with placeholders for entities. We align these placeholders too with 2b2 medical entity types.  We use a subset of the schema in our LFs, as summarized in Fig.\ref{sem}, to represent the relation between the entities in the question and the answer.}  

A formal representation of the LF grammar using this schema (Figure \ref{sem}) is as follows. Medical events are denoted as $ME_{i}$ (e.g LabEvent, ConditionEvent) and  relations are denoted as $RE_{i}$ (e.g conducted/reveals). Now, $ME[a_{1},..,a_{j},..,oper(a_{n})]$ is a medical event where $a_{j}$ represents the attribute of the event (such as result in LabEvent). An event may optionally include constraints on attributes captured by an operator ($oper()$ $\in$ sort, range, check for null values, compare).  These operators sometimes require values from external medical KB (indicated by \textit{ref}, e.g. lab.\textit{ref}low/lab.\textit{ref}high to indicate range of reference standards considered healthy in lab results) indicating the need for medical knowledge to answer the question. Using these constructs, a $LF$ can be defined using the following rules,\\
$LF \rightarrow ME_{i}$ $|$ $M_{1}$ relation $M_{2}$\\
$M_{1} \rightarrow ME_{i}$, $M_{2} \rightarrow ME_{j}$\\
$M_{1} \rightarrow M_{1}$ relation $M_{2}$, 
$M_{2} \rightarrow M_{1}$ relation $M_{2}$\\
$relation \rightarrow$ $OR$ $|$ $AND$ $|$ $RE_{i}$
%In addition we also define medical knowledge base entities corresponding to each medical event - the properties of the KB entity capture general medical knowledge about that entity. 
%Logical forms $LF$ capture the the different relations $RE_{i}$ between the entities $ME_{i}$ in the question templates and the answer entities $ME_{j}$, and help us formalize question characteristics and also collect answers (section \ref{answers}). 
% The events and attributes accept as arguments the different entities in the clinial note to    
% In Table \ref{qlfo} \todo{JL comments on Table 1: (1) "Is the test value high" - do we need to explain why "date" is an answer type even though the question doesn't ask for a date?}, 
% we show that logical forms express single, multiple and compositional relation patterns between the event entities in the question the answer. The questions with no relation in the question are either the questions about event attributes or open-ended questions (e.g. mention of $|medication|$ ?).

%The manual process of annotating logical forms based on the ontology provides a way for us to explicitly capture the specific answer of physicians interest.
Advantages of our LF representation include the ability to represent composite relations, define attributes for medical events and constrain the attributes to precisely capture the information need in the question. While these can be achieved using different methods that combine lambda calculus and first order logic \cite{roberts2016annotating}, our representation is more human comprehensible. This allows a physician to consider an ontology like Figure \ref{sem} and easily  define a logical form. Some example question templates with their $LF$ annotations are described in Table \ref{tab:55} using the above notation. The $LF$ representation of the question in Figure \ref{fig:mesh1} is \textit{MedicationEvent($|medication|$) [dosage=x]}. The entities seen in $LF$ are the entities posed in the question and entity marked $x$ indicates the answer entity type.

\begin{table*}[t]
\centering
\footnotesize
  \begin{tabular}{p{4cm} @{\hspace{0.1\tabcolsep}} p{11cm} @{\hspace{0.1\tabcolsep}}p{0.5cm}}
  \hline
 Property & Example Annotation & Stats. \\
 \hline
 %Avg. \# of entity types per question & Does the patient take \textbf{mode} \textbf{medication} for \textbf{problem} & 1.5  \\
 %47.9,52.2,18.1,10.6,46.8
 Fine grained answer type \newline (attribute entity is answer) & Q: What is the \textbf{dosage} of $|medication|$ ? \newline LF: MedicationEvent ($|medication|$) [\textbf{dosage=x}]  & 62.7\%   \\ \hline
 Course grained answer type \newline (event entity is answer) & Q: \textbf{What} does the patient take $|medication|$ for? \newline LF: MedicationEvent($|medication|$)given\{ConditionEvent(\textbf{x}) OR SymptomEvent(\textbf{x})\}& 52.1\% \\ \hline
Questions with operators on \newline entities   & Q: What are the last set of labs with \textbf{elevated} numbers out of range? \newline LF: LabEvent (x) [date=x, (result=x)$\boldsymbol{>}$lab.refhigh]  & 25.5\%   \\  \hline
Questions which require \newline medical KB   &  Q: What are the last set of labs with \textbf{elevated numbers out of range}? \newline LF: LabEvent (x) [date=x, (result=x)$>$\textbf{lab.refhigh}]   & 11.7\%   \\  \hline
% Descriptive (questions looking for justification) & Q: \textbf{Has} patient ever been prescribed $|medication|$? \newline LF: MedicationEvent ($|medication|$) & 50.3\% \\
% \hline
At least one event relation   &  What lab results does he have that are pertinent to $|problem|$ \textbf{diagnosis} \newline LF: LabEvent (x) [date=x, result=x] \textbf{conducted/reveals} ConditionEvent ($|problem|$) & 46.8\%   \\
\hline
  \end{tabular}
  \caption{Properties of question templates inferred from the corresponding logical form templates. The boldface words hint at the presence of the corresponding property in both question and the logical form template.}
  \label{tab:55}
\vspace{-0.1in}
\end{table*}
%\todo{you can merge this with the lines marked red in answer extraction, but use this to give a justification of how manual annotations helped us which otherwise would have been miselead just by normal interpretation. }
%LFs are manually annotated by physicians who use it to capture the answer they are looking for. For e.g, \textit{Does the patient have any medication allergies} would normally be interpreted as a \textit{yes/no} question, but the annotated LF \textit{MedicationEvent (x) causes {ConditionEvent (x) OR SymptomEvent (x)}} explicitly shows that the physician is interested in a list of all the medications and conditions or symptoms caused by it. 
%Using this information, we build a rule based framework that extracts existing annotations containing entities that satisfy the relations expressed in the LF.

%(Table  \ref{para-table}). %\todo{Can you say how many pairs of question paraphrases we have?} 
%\todo{make a connection between this formalism and the example LFs in the table. This seems a bit disconnected now.}
%\vspace{-0.12in}
\subsection{Template Filling and Answer Extraction}
\label{answers}
%\vspace{-0.06in}
The next step in the process is to populate the question and logical form (QL) templates with existing annotations in the i2b2 clinical  datasets and extract answer evidence for the questions. 
%The entity placeholders in the questions and logical form templates and the answers are filled using i2b2  annotations. 
%\textbf{I2b2 Challenge Datasets:}
The i2b2 datasets are expert annotated with fine-grained annotations \cite{guo2006identifying} that were developed for various shared NLP challenge tasks, including (1) smoking status classification \cite{uzuner2008identifying}, (2) diagnosis of obesity and its co-morbidities \cite{uzuner2009recognizing}, extraction of  (3) medication concepts \cite{uzuner2010extracting}, (4) relations, concepts, assertions \cite{uzuner2010community, uzuner20112010} (5) co-reference resolution \cite{uzuner2012evaluating} and (6) heart disease risk factor identification \cite{stubbs2015annotating}. In Figure \ref{fig:mesh1}, this would correspond to leveraging annotations from medications challenge between medications and their dosages, such as \textit{medication=Nitroglycerin, dosage=40mg},  to populate $|$medication$|$ and generate several instances of the question ``What is the dosage of $|medication|$?" and its corresponding logical form \textit{MedicationEvent($|medication|$)[dosage=x]}. The answer would be derived from the value of the dosage entity in the dataset. 

%The annotated entities in these datasets are used to fill placeholders in our QL pair templates, whereas the appropriate relations expressed in (4),(6) and attributes in (3) are mapped to the relations in templates \cm{where entities are linked by a relation}.
%In summary, the i2b2 annotations provide relation patterns between different event entities or event-attribute values in the EMR. 

%i2b2 annotation guidleines
%\textbf{Preprocessing:} These annotations are preprocessed before using them with our templates to ensure syntactic correctness of the generated questions. The processing steps includes excluding common nouns with the help of WordNet \cite{miller1995wordnet} \footnote{We consider the top-300 high frequency nouns occurring in our question templates as common nouns and expand them using wordnet to 742 common-nouns}, stopword removal, replacing possessive nouns, replacing annotations that start with a determiner, preposition, subordinating conjunction or wh-determiner with an empty string. To estimate accuracy, we randomly sample 500 generated questions, where observed grammatical errors are $<$5\%  

\textbf{Preprocessing:} The i2b2 entities are preprocessed before using them with our templates to ensure syntactic correctness of the generated questions. The pre-processing steps are designed based on the i2b2 annotations syntax guidelines \cite{guo2006identifying}. To estimate grammatical correctness, we randomly sampled 500 generated questions and found that $<$5\% had errors. These errors include, among others, incorrect usage of article with the entity and incorrect entity phrasing.

\textbf{Answer Extraction:} 
%\todo{from what},%ConditionEvent (|problem|) [status=x] OR SymptomEvent (|problem|) [status=x],
The final step in the process is generating answer evidence corresponding to each question. The answers in emrQA are defined differently; instead of a single word or phrase we provide the entire i2b2 annotation line from the clinical note as the answer. This is because the context in which the answer entity or phrase is mentioned is extremely important in clinical decision making \cite{demner09}. Hence, we call them answer evidence instead of just answers.  For example, consider the question \textit{Is the patient's $hypertension$ controlled?}. The answer to this question is not a simple \textit{yes/no} since the status of the patient's $hypertension$ can change through the course of treatment. The answer evidence to this question in emrQA are multiple lines across the longitudinal notes that reflect this potentially changing status of the patients condition, e.g. \textit{Hypertension-borderline today}. Additionally, for questions seeking specific answers we also provide the corresponding answer entities.
% In emrQA, we have questions seeking specific answers like \emph{What is the dosage of $|medication|$?} where the answer to the question is an entity of type \emph{$dosage$} and \cm{other questions like``\emph{Does the }", where the answer to this question is not a simple \textit{yes/no}. 
% since patient's $medication$ usage can change through the course of treatment e.g. \textit{No Insulin today}.}

% \cm{This difference can be learned from the corresponding annotated logical forms, that express the physicians answer need.} %expressing what the physician is looking for to answer a question.} 

% So in both the cases, we provide answer evidence, which is the entire i2b2 annotation line from the clinical note. 
%This is because the context in which the answer entity or phrase is mentioned is extremely important to the physician \cite{demner09}. 

%The answers in emrQA are defined differently, instead of a single word we provide the entire i2b2 annotation line from the clinical note as the answer. 
% For e.g., .  The answer evidence for this in emrQA is multiple lines across longitudinal notes that may reflect the changing status of the patients problem, .
%Similarly, consider the question, \emph{What is the dosage of |medication|?}, 
The overall process for answer evidence generation was vetted by a physician. Here is a brief overview of how the different i2b2 datasets were used in generating answers.
%one paragraph explaining medication co-reference  not used limitation
The \textit{relations challenge} datasets have various event-relation annotations across single/multiple lines in a clinical note. We used a combination of one or more of these, to generate answers for a question; in doing so we used the annotations provided by the \textit{i2b2 co-reference} datasets. Similarly, the \textit{medications challenge} dataset has various event-attribute annotations but since this dataset is not provided with co-reference annotations, it is currently not possible to combine all valid answers.
%\todo{What do you want to say here}
%Hence some of these questions will require multiple sentence reasoning to gather answer evidence.
%\todo{define longitudinal somewhere, preferably at the first usage}
The \textit{heart disease challenge dataset} has longitudinal notes ($\sim$5 per patient) with record dates. The events in this dataset are also provided with time annotations and are rich in quantitative entities. This dataset was primarily used to answer questions that require temporal and arithmetic reasoning on events. %\todo{check for event-entity usage}
%Also the entity annotations provided on the records of this dataset are noisy and unfit to use with the question templates, so instead we use the standard entity names \footnote{also provided with the i2b2-annotations} of these entities in the questions and return the evidence from the note \todo{what does this mean?}. This results in a greater lexical variation between the question sentence and answer evidence.
The  patient records in the \textit{smoking and obesity challenge} datasets are categorized into classes with no entity annotations. Thus, for questions generated on these datasets, the entire document acts as evidence and the annotated class information (7 classes) needs to be predicted as the answer.

The total questions, LFs and answers generated using this framework are summarized in Table \ref{tab:3}. Consider the question \textit{How much does the patient smoke?} for which we do not have i2b2-annotations to provide an answer. In cases where the answer entity is empty, we only generate the question and LF, resulting in more question types being used for QL than QA pairs: only 53\% of question types have answers. 

%% file: emrqa_analysis.tex
\vspace{-0.05in}
\section{emrQA Dataset Analysis}
\label{analysis}
\vspace{-0.04in}
%\vspace{-0.06in}
We analyze the complexity of emrQA by considering the LFs for question characteristics, variations in paraphrases, and the type of reasoning required for answering questions (Table \ref{para-table}, \ref{tab:55}, \ref{reasoning-categories}). 
\subsection{Question/Logical Form Characteristics} % dont know why making this a subsection
\label{qlchar}
%\vspace{-0.06in}
A  quantitative  and  qualitative  analysis of emrQA question templates is shown in Table \ref{tab:55}, where logical forms help formalize their characteristics \cite{su2016generating}.  Questions may request specific fine-grained information (attribute values like dosage) or may express a more coarse-grained need (event entities like medications etc), or a combination of both. 25\% of questions  require complex operators (e.g compare($>$)) and  12\% of questions express the need for external medical knowledge (e.g. lab.refhigh).  The questions in emrQA are highly compositional, where 47\% of question templates have at least one event relation.
 %The distribution of emrQA and emrQL question properties on using them with i2b2 annotations is also summarized in \tabref{tab:55}
%Avg. entity type per question & When did the patient have (test/problem/treatment) ? & 1.014 \\
% Arithmetic Reasoning & Show me any $|test|$ $\mathbf{>}$ $|value|$ in the \textbf{last} $|time|$ & 26.59 \%  \\ 
%Medical Reasoning &  Is the $|test|$ value \textbf{abnormal} & 12.76 \% \\
%\vspace{-0.08in}
\subsection{Paraphrase Complexity Analysis}
%\vspace{-0.06in}
%sentence tokenizer gave 1.076 >1 implies there are cases where the question had 2 sentences
%emrQA covers various topics of physician interest.
%Q: What were the results of the abnormal \textbf{blood pressure} in 2084-03-05 \newline E: \textbf{BP} 170/100\\
%Inference &  Sentences which will need non medical specific inference & Q: Did the patient \textbf{receive} diuresis for a lower extremity edema  \newline E: His lower extremity edema \textbf{improved} with diuresis \\
\begin{table*}[t]
\centering
\footnotesize
\renewcommand{\arraystretch}{1.1}
  \begin{tabular}{@{\hspace{-0.1\tabcolsep}} p{2.5cm} @{\hspace{0.9\tabcolsep}} p{4.5cm} @{\hspace{1.0\tabcolsep}} p{6.4cm} @{\hspace{0.6\tabcolsep}} c @{\hspace{0.6\tabcolsep}} c }
  \hline
Reasoning & Description   &  Example Annotation &  emrQA & SQuAD \\
 \hline
Lexical Variation \newline (Synonym) &   Major correspondence between the question and answer sentence are synonyms. & Q: Has this patient ever been \textbf{treated} with insulin? \newline E: Patient sugars were \textbf{managed} o/n with sliding scale insulin and diabetic & 15.2\% & 33.3\%\\
\hline
Lexical Variation \newline (world/medical knowledge)  &  Major correspondence between the question and answer sentence requires world/medical knowledge to resolve&Q: Has the patient complained of any \textbf{CAD symptoms}? \newline E:  70-year-old female who comes in with \newline \textit{\textbf{substernal chest pressure}} & 39.0\% & 9.1\%\\
\hline
Syntactic Variation  &  After the question is paraphrased
into declarative form, its syntactic
dependency structure does not
match that of the answer sentence
 & Q: Has this patient ever been \textbf{treated} with ffp?  \newline E: attempt to reverse anticoagulation , one unit of \textbf{FFP was begun}  & 60.0\% &64.1\% \\
\hline
Multiple Sentence & Co-reference and higher level fusion of multiple sentences & Q: What happened when the patient was given \textbf{ascending aortic root replacement}? \newline E: The patient tolerated \textbf{the procedure} fairly well and was transferred to the ICU with his \textit{chest open} & 23.8\% & 13.6\% \\
\hline
Arithmetic& Knowing comparison and subtraction operators.  &Q:  Show me any \textbf{LDL $>$ 100 mg/dl} in the last 6 years? \newline E: gluc 192, \textit{\textbf{LDL 115}}, TG 71, HDL 36  & 13.3\% & N.A.\\
\hline
Temporal &  Reasoning based on time frame  &Q: What were the results of the abnormal A1C on \textbf{2115-12-14}? \newline E: HBA1C \textbf{12/14/2115} \textit{11.80} & 18.1\%  & N.A. \\
\hline
Incomplete \newline Context & Unstructured clinical text is noisy and may have missing context  & Q: What is her current dose of iron?
\newline E: Iron \textit{325 mg} p.o. t.i.d. & 28.6\%  & N.A.  \\
\hline
Class Prediction & Questions for which a specific predefined class needs to be predicted  &  Q: Is the patient currently Obese? \newline E: Yes &  12.4\%  & N.A. \\
\hline
  \end{tabular}
  \caption{We manually labeled 105 examples into one or more of the above categories. Words relevant to the corresponding
reasoning type are in bold and the answer entity (if any) in the evidence is in $italics$. We compare this analysis with SQuAD.}
%\vspace{-0.2in}
  \label{reasoning-categories}
\end{table*}

Questions templates that map to the same LF are considered paraphrases (e.g, Table \ref{para-table}) and correspond to the same question type.
%since they express the same information need they belong 
%map to the same question type. 
%Some example question paraphrases are shown in Tab \ref{para-table}.  
In emrQA, an average of 7 paraphrase templates exist per question type. This is representative of FAQ types that are perhaps more important to the physician. Good paraphrases are lexically dissimilar to each other \cite{chen2011collecting}. In order to understand the lexical variation within our paraphrases, we randomly select a question from the list of paraphrases as a reference and evaluate the others with respect to the reference, and report the average BLEU (0.74 $\pm$ 0.06) and Jaccard Score (0.72 $\pm$ 0.19).  The low BLEU and Jaccard score with large standard deviation indicates the lexical diversity captured by emrQA's paraphrases  \cite{papineni2002bleu, niwattanakul2013using}.
%\0.06
%Report TER
%percentage of question with paraphrases
%In order to show the diversity in the words used in our paraphrases we show the range of pairwise Jaccard Index in our question paraphrases. The scores range from 0 to 1. Low score implies large difference in the words used in the pair of question paraphrases. Large number of paraphrases (7.2 per question type)  indicates the question types frequently asked and consequently perhaps more important to the physician.
%emrQL has more paraphrase complexity compared to others, %FLES score is lower than all of them, which shows questions difficulty.% emrQL has less number of types compared to GeoQUery, NLMaps and Freebase because it is more closed domain in nature compared to these datasets, and lack of annotations to build them currently.
%\vspace{-0.1in}
\subsection{Answer Evidence Analysis}
%consider one paraphrase question at random
%questions which dont have answer pairs because of missing i2b2 annotations
%not he entire dataset
\label{evidence-analysis}
%\vspace{-0.06in}
33\% of the questions in emrQA have more than one answer evidence, with the number ranging from 2 to 61. E.g., the question \textit{Medications Record?} has all medications in the patient's longitudinal record as answer evidence. In order to analyze the reasoning required to answer emrQA questions, we sampled 35 clinical notes from the corpus and analyzed 3 random questions per note by manually labeling them with the categories described in Table \ref{reasoning-categories}.  Categories are not mutually exclusive: a single example can fall into multiple categories. We compare and contrast this analysis with SQuAD  \cite{rajpurkar2016squad}, a popular MC dataset generated through crowdsourcing, to show that the framework is capable of generating a corpus as representative and even more complex. Compared to SQuAD, emrQA offers two new reasoning categories, temporal and arithmetic which make up 31\% of the dataset. Additionally, over two times as many questions in emrQA require reasoning over multiple sentences. Long and noisy documents make the question answering task more difficult \cite{joshi2017triviaqa}. EMRs are inherently noisy and hence 29\% have incomplete context and the document length is 27 times more than SQuAD which offers new challenges to existing QA models. Owing to the domain specific nature of the task, 39\% of the examples required some form of medical/world knowledge.

As discussed in Section \ref{answers}, 12\% of the questions in emrQA corpus  require a class category from \textit{i2b2 smoking and obesity datasets} to be predicted. We also found 6\% of the questions had other possible answers that were not included by emrQA, this is because of the lack of co-reference annotations for the \textit{medications challenge}.
%(3761, 140.63)
%were valid answers that were not included i
%and as discussed in section \ref{answers}, large part of this contribution comes from the i2b2's heart disease risk dataset.
% Based on the QA pairs generation process and the nature of i2b2 annotations, we can roughly understand the reasoning required per dataset, which is summarized in \ref{reasoning-dataset-1}. 
%QA pairs generated from risk data analysis will be rich in (a) lexical variation (both synonym and world knowledge). This is because the concepts used in the question are not extracted from the clinical note.  (b) The medical reasoning involved to answer these questions is high given the nature of the dataset 
%QA pairs generated from the relations and medications dataset can be answered from the context given in the notes. (a) Low on Lexical Variation (both synonym and world knowledge) since concepts used in the question come from clinical notes. (b) They will not require medical reasoning the information can need to be inferred from the note itself. 
%The document classification task in smoking and obesity challenge, will have lexical variation (synonym and world), high level inference of class information, sentences which will need multiple reasoning. No arithmetic and temporal reasoning.
%\vspace{-0.13in}

%% file: emrqa_baselines.tex
\vspace{-0.05in}
\section{Baseline Methods}
\vspace{-0.04in}
\label{baselines}
%\vspace{-0.1in}
%To quantify the difficulty level of the dataset for current methods, we present results on simple heuristic models. We provide methods for both Q-A and Q-L datasets and discuss the utility of both the datasets to encourage research in both directions. 
%We implement baseline models for both question - logical (Q-L) form and question - answer (Q-A) mapping.  
%In this section, we discuss heuristic model for Q-L mapping that will help us understand the implications of using emrQA for semantic parsing of questions. This will also help us understand the problems or steps that a conventional neural or machine learning model will face.
We implement baseline models using neural and
heuristic methods for question to logical form (Q-L) and question to answer (Q-A) mapping.

%\vspace{-0.08in}
\subsection{Q-L Mapping}
%\vspace{-0.1cm}
%challenging compared to just the entity-level paraphrases.
%with  GeoQuery \cite{zelle1996learning} and ATIS \cite{dahl1994expanding}.
%We do so to understand if our paraphrases are learnable.
%dont know if entity and sentence kevel paraphrases term makes sense
%template based approach 
%is the \cm{inverse of the generation process}
%For learning logical forms from question, we implement heuristic models and a seq2seq  neural baseline \cite{luong17}.
\label{qlmodels}
\textbf{Heuristic Models:} We use a template-matching approach where we first split the data into train/test sets, and then normalize questions in the test set into templates by replacing entities with placeholders. The templates are then scored against the ground truth templates of the questions in the train set, to find the best match. The placeholders in the LF template corresponding to the best matched question template is then filled with the normalized entities to obtain the predicted LF. To normalize the test questions we use CLiNER \cite{boag2015cliner} for emrQA and \newcite{jia2016data}'s work for ATIS and GeoQuery. Scoring and matching is done using two heuristics: (1) HM-1, which computes an identical match, and (2) HM-2, which  generates a GloVe vector \cite{arora2016simple} representation of the templates using sentence2vec  and then computes pairwise cosine similarity.
%and GeoQuery, ATIS questions using \cite{jia2016data} work.
%Even though the data is generated from templates, the entities added to the QL pair templates are complex and diverse to make the inverse process non trivial. Thus resulting in a 
%\vspace{-0.1in}
\begin{table}[t]
\footnotesize
\centering
\footnotesize
\begin{tabular}{@{\hspace{-0.1\tabcolsep}}l|l|l|l|l@{\hspace{-0.1\tabcolsep}}}
\hline
\textbf{Dataset}   &  \textbf{Train/Test} & \textbf{HM-1} & \textbf{HM-2} & \textbf{Neural} \\ \hline
GeoQuery   & 600/280 &32.8\%            & 52.1\%  &   74.6\%\tablefootnote{\label{note1}results from \newcite{jia2016data}}       \\ 
ATIS      & 4,473/448  &20.8\%            & 52.2\%  &   69.9\%\textsuperscript{\ref{note1}}      \\ 
emrQL-1      & 1M/253K  & 0.3\%             & 26.3\% &   22.4\%        \\ 
emrQL-2       &  1.1M/296K  & 31.6\%            &  32.0\% &   42.7\%         \\ \hline
\end{tabular}
\caption{Heuristic (HM) and neural (seq2seq) models  performance on question to logical form learning in emrQA.}
%\vspace{-1.2cm}
\vspace{-0.2in}
\label{semantic-parsing}
\end{table}
%\vspace{-0.1in}

\textbf{Neural Model:} We train a sequence-to-sequence (seq2seq) \cite{sutskever2014sequence} with attention paradigm \cite{bahdanau2014neural, luong17} 
%using the implementation by \cite{luong17})  
as our neural baseline (2 layers, each with 64 hidden units). The same setting when used with Geoquery and ATIS gives poor results because the parameters are not appropriate for the nature of that dataset. Hence, for comparison with GeoQuery and ATIS, we use the results of seq2seq model with a single 200 hidden units layer \cite{jia2016data}. At test time we automatically balance missing right parentheses.

%We train a neural baseline we use an existing TensorFlow model by \cite{luong17} based on
%\vspace{-0.12in}
\subsubsection{Experimental Setup}
%\vspace{-0.06in}
%Number of elements in each split ?
We randomly partition the QL pairs in the dataset in train(80\%) and test(20\%) sets in two ways. (1) In emrQL-1, we first split the paraphrase templates corresponding to a single LF template into train and test, and then generate the instances of QL pairs. (2) In emrQL-2, we first generate the instances of QL pairs from the templates and then distribute them into train and test sets. As a result, emrQL-1 has more lexical variation between  train and test distribution compared to emrQL-2,  resulting in increased paraphrase complexity. We use accuracy i.e, the total number of logical forms predicted correctly as a metric to evaluate our model.

%Error analysis of our baselines shows that QL mapping task of emrQA dataset is challenging and supports significant future work.

% \begin{figure}[t]
% \begin{floatrow}
% \footnotesize
% \hspace{-0.5cm}
% \ffigbox{%
%   \includegraphics[width=0.5\textwidth]{images/error_ql}
% }{%
%   \caption{Error Analysis of Seq2Seq}%
% }
% \hspace{-1.0cm}
% \capbtabbox{%
%   \begin{tabular}{c|c|c}
%   \hline
% \textbf{Model}  & \textbf{EM} & \textbf{F1}  \\ \hline
% MC (Overall)  & 59.2\% & 60.6\\
% MC (Descriptive)   & 59.6\% & 63.3 \\
% MC (Specific)  &  58.8\% &  57.9 \\
% Class Prediction  & 21.1\% & n.a \\
% \hline
% \end{tabular}%
% }{%
%   \caption{Performance of baseline models on two of its sub tasks, machine comprehension and }%
% }
% \end{floatrow}
% \end{figure}

%\vspace{-0.09in}
\subsubsection{Results}
%\vspace{-0.05in}
The performance of the proposed models is summarized
in Table \ref{semantic-parsing}. emrQL results are not directly comparable with GeoQuery and ATIS because of the differences in the lexicon and tools available for the domains. However, it helps us establish that QL learning in emrQA is non-trivial and supports significant future work.
%Error analysis revealed that the accuracy of heuristic models on emrQL-1 and emrQL-2 is low because 70\% errors are caused at the normalization step (sec \ref{qlmodels}), where 30\% of the questions have not been normalized. 

Error analysis of heuristic models on emrQL-1 and emrQL-2 showed that 70\% of the errors occurred because of incorrect question normalization. In fact, 30\% of these questions had not been normalized at all. This shows that the entities added  to the templates are complex and diverse and make the inverse process of template generation non trivial. This makes a challenging QL corpus that cannot trivially be solved by template matching based approaches.

\begin{figure*}[ht!]
\footnotesize
\centering
  \begin{tabularx}{2\linewidth}[t]{*{2}X}
   % \hline
     \begin{tabular}{@{\hspace{-0.1\tabcolsep}} p{10cm}@{\hspace{0.0001\tabcolsep}}  | p{3cm} @{\hspace{0.1\tabcolsep}}  | p{2cm} }
  \hline
 \textbf{Logical Form template}  & \textbf{Property}  & \textbf{Exact Match}   \\ \hline
% medical KB &\textit{LabEvent (x) [date=x, (result=x)\textbf{$>$}\textbf{lab.refhigh}]}       & 0.0       \\ 
%   \hline
% arithmetic operators &\textit{LabEvent ($|test|$) [(date=x)$>$(currentDate-$|time|$), \textbf{(result=x)$>$$|value|$}]}  &  0.0   \\ 
%   \hline
%  MedicationEvent (x) & descriptive &  10.9\% \\ \hline
 MedicationEvent ($|medication|$) [\textbf{enddate=x}] & single attribute & 55.3\% \\ \hline
 \{LabEvent ($|test|$) OR ProcedureEvent ($|test|$)\} \textbf{conducted}\newline \{ConditionEvent(x) OR SymptomEvent (x)\} & single  relation & 32.2\% \\ \hline
 \{MedicationEvent($|treatment|$)ORProcedureEvent($|treatment|$)\} \newline \textbf{improves/worsens/causes} \{ConditionEvent (x) OR SymptomEvent (x)\} & multiple  relation & 12.6\% \\ \hline
%course grained answer &\textit{MedicationEvent ($|medication|$) [\textbf{sig=x}]} & 87.4\% \\
    \end{tabular} &
    \centering
 \tabularnewline
 
    \captionof{table}{Neural models (DrQA) performance on question-evidence corpus of emrQA stratified according to the logical form templates. Instance showing increasing complexity in the logical forms with decreasing model performance.}\label{drqa-error} &
     \tabularnewline
\end{tabularx}
\vspace{-1.2cm}
\end{figure*}
Errors made by the neural model on both emrQL-1 and emrQL-2 are due to long LFs (20\%) and incorrectly identified entities (10\%), which are harder for the attention-based model \cite{jia2016data}. The increased paraphrase complexity in emrQL-1 compared to emrQL-2 resulted in 20\% more structural errors in emrQL-1, where the predicted event/grammar structure deviates significantly from the ground truth. This shows that the model is not adequately capturing the semantics in the questions to generalize to new paraphrases. Therefore, emrQL-1 can be used to benchmark QL models robust to paraphrasing.

\subsection{Q-A Mapping}
%\vspace{-0.06in}
Question-answering on emrQA consists of two different tasks, (1) extraction of  answer line from the clinical note (machine comprehension (MC)) and (2) prediction of answer class based on the entire clinical note. We provide baseline models to illustrate the complexity in doing both these tasks.

\textbf{Machine Comprehension}: To do extractive QA on EMRs, we use DrQA's \cite{chen2017reading} document reader which is a multi-layer RNN based MC model. We use their best performing settings trained for SQuAD data using Glove vectors (300 dim-840B). 
%Number of classes and what are they ?

\textbf{Class Prediction:} We build a multi-class logistic regression model for predicting a class as an answer based on the patient's clinical note. Features input to the classifier are TF-IDF vectors of the question and the clinical notes taken from \textit{i2b2 smoking and obesity datasets}. 
% \begin{table*}[t]
% \footnotesize
% \centering
%   \begin{tabular}{@{\hspace{-0.1\tabcolsep}} p{9.3cm}@{\hspace{0.0001\tabcolsep}} | p{2cm} @{\hspace{0.1\tabcolsep}}| p{1cm}}
%   \hline
%  \textbf{Logical Form template}  & \textbf{Property}  & \textbf{EM}   \\ \hline
% % medical KB &\textit{LabEvent (x) [date=x, (result=x)\textbf{$>$}\textbf{lab.refhigh}]}       & 0.0       \\ 
% %   \hline
% % arithmetic operators &\textit{LabEvent ($|test|$) [(date=x)$>$(currentDate-$|time|$), \textbf{(result=x)$>$$|value|$}]}  &  0.0   \\ 
% %   \hline
% %  MedicationEvent (x) & descriptive &  10.9\% \\ \hline
% \{MedicationEvent($|treatment|$)ORProcedureEvent($|treatment|$)\} \newline \textbf{improves/worsens/causes} \{ConditionEvent (x) OR SymptomEvent (x)\} & multiple \newline relation & 12.6\% \\ \hline
%   \{LabEvent ($|test|$) OR ProcedureEvent ($|test|$)\} \textbf{conducted}\newline \{ConditionEvent(x) OR SymptomEvent (x)\} & single \newline relation & 32.2\% \\ \hline
%  MedicationEvent ($|medication|$) [\textbf{enddate=x}] & single attribute & 55.3\% \\ \hline
% %course grained answer &\textit{MedicationEvent ($|medication|$) [\textbf{sig=x}]} & 87.4\% \\
% \end{tabular}%
% \caption{Neural models performance on question-evidence corpus of emrQA stratified according to the logical form templates. }
% \label{map_dataset}
%  \vspace{-0.20in}
% \end{table*}
%\vspace{-0.13in}
\subsubsection{Experimental setup}
%\vspace{-0.05in}
% In the MC model setting, we build our question-evidence pair on any given clinical note by randomly selecting a question from the list of our paraphrases.  This is a more conserved setting indented to analyze the worst case performing model. In a more informed setting all the paraphrases of a question can be used  to train with an evidence in the clinical note. We then randomly divide our clinical notes into 80\% train and 20\% test. We use a slightly modified version of the two popularly reported metrics in MC to evaluate our models, Exact Match (EM) and F1. Wherever the answer entity in an evidence is explicitly given, EM measures if the entity is present within the evidence, otherwise we check if the  predicted evidence span lies within $\pm$ 20 characters of the ground truth evidence. For F1 match we first construct a bag of tokens for each evidence and measure the F1 score of the overlap between the two bags of tokens. Since we have multiple evidence for a given question, we consider the top 10 predictions and report an average EM and F1 over ground truth number of answers. In the class prediction setting, we report the subset accuracy over the 7 classes taken from smoking and obesity datasets.
We consider a 80-20 split of the data for train-test. In order to evaluate worst-case performance, we train on question-evidence pairs in a clinical note obtained by using only one random paraphrase for a question instead of all the paraphrases. We use a slightly modified\footnote{using the original definitions, the evaluated values were far less than those obtained in Table \ref{drqa_results}} version of the two popularly reported metrics in MC for evaluation since our evidence span is longer: Exact Match (EM) and F1. Wherever the answer entity in an evidence is explicitly known, EM checks if the answer entity is present within the evidence, otherwise it checks if the predicted evidence span lies within $\pm 20$ characters of the ground truth evidence. For F1  we construct a bag of tokens for each evidence string and measure the F1 score of the overlap between the two bags of tokens. Since there may be multiple evidence for a given question, we consider only the top 10 predictions and report an  average of EM and F1 over ground truth number of answers. In the class prediction setting, we report the subset accuracy.
\begin{table}[t]
\footnotesize
  \begin{tabular}{c|c|c|c}
  \hline
\textbf{Model} & \textbf{Train/Test} &\textbf{Exact Match} & \textbf{F1}  \\ \hline
DrQA (MC)  & 47,605/9,966 & 59.2\% & 60.6\\
Class Prediction & 1276/320 & 36.6\% & n.a \\
\hline
\end{tabular}%
\caption{Performance of baseline models on the two QA sub tasks, machine comprehension (MC) and class prediction.}
\label{drqa_results}
\vspace{-0.2in}
\end{table}
% \begin{table}[t]
% \footnotesize
% \centering
%   \begin{tabular}{c|c|c|c}
%   \hline
% \textbf{Model} & \textbf{Train/Test} & \textbf{EM} & \textbf{F1}  \\ \hline
% MC (Overall) & 47,605/9966 & 59.2\% & 60.6\%\\
% MC (Descriptive)  & 23,979/4965 & 59.6\% & 63.3\% \\
% MC (Specific) & 23,626/5001 &  58.8\% &  57.9\% \\
% Class Prediction & 894/224 & 28.5\% & n.a \\
% \hline
% \end{tabular}%

% ('EM descptive', 0.47616649882510914)
% ('EM specfic', 0.5881407845415043)
% ('EM total', 0.5323558830181346)
% ('F1 descptive', 0.6335654800744466)
% ('F1 specfic', 0.5793000783312351)
% ('F1 total', 0.6063347682424389)

% \begin{table}[t]
% \footnotesize
% \centering
%   \begin{tabular}{c|c}
%   \hline
% \textbf{Category} & \textbf{Percentage}  \\ \hline
% Multiple Sentence       &       \\ 
% Arithmetic Reasoning      &     \\ 
% Temporal Reasoning         & \\
% \hline
% \end{tabular}%
% \caption{Performance of emrQA and per i2b2-dataset generated QA pairs measured by considering Macro average of MAP (Mean Average Precision), shows that heart disease risk factors dataset is the most challenging. }
% \label{map_dataset}
%  \vspace{-0.20in}
% \end{table}
%In order to avoid bias caused by the nature of i2b2 dataset used for generation, we first observe only the questions that belong to the medications challenge. 
%, using their metric emrQA evaluates to F1 50.8.

%\vspace{-0.06in}
\subsubsection{Results}
%\vspace{-0.03in}
The performance of the proposed models is summarized in Table \ref{drqa_results}. DrQA is one of the best performing models on SQuAD with an F1 of 78.8 and EM of 69.5. The relatively low performance of the models on emrQA  (60.6 F1 and 59.2 EM) shows that QA on EMRs is a complex task and offers new challenges to  existing QA models.

%In order to qualitatively and quantitatively understand the errors done by the MC model, authors typically try to understand the underlying reasoning involved to answer the question. They do so by a statistical process similar to the one in Sec.\ref{evidence-analysis}. 
To understand model performance, we macro-average the EM across all the questions corresponding to a LF template. We observe that LFs representing temporal and arithmetic\footnote{\label{hdr}maximum representation of these templates comes from the i2b2 heart disease risk dataset} needs had $<16\%$ EM.  LFs expressing the need for medical KB\textsuperscript{\ref{hdr}} performed poorly since we used general Glove embeddings. An analysis of LFs which had approximately equal number of QA pair representation in the test set revealed an interesting relation between the model performance and LF complexity, as summarized in Table \ref{drqa-error}. The trend shows that performance is worse on multiple relation questions as compared to single relation and attribute questions, showing that the LFs sufficiently capture the complexity of the questions and give us an ability to do a qualitative model analysis. 

Error analysis on a random sample of 50 questions containing at least one answer entity in an evidence showed that: (1) 38\% of the examples required multiple sentence reasoning of which 16\% were due to a missing evidence in a multiple evidence question, (2) 14\% were due to syntactic variation, (3) 10\% required medical reasoning and (4) in 14\%, DrQA predicted an incomplete evidence span missing the answer entity in it. 

%% file: discussion.tex
\section{Discussion}
\vspace{-0.04in}
%Our proposed frameworks ability to generate a large scale dataset on any complex domain such as EMRs is demonstrated in Section \ref{emrqa}. In Section \ref{analysis} and \ref{baselines}, by analyzing emrQAs dataset complexity we also established the frameworks potential to generate a challenging dataset. 
%
%In this section, we attempt to widen the outlook of the frameworks application capability by briefly outlining how the process may be applied to any open-domain QA dataset generation.-->
In this section, we describe how our generation framework may also be applied to generate open-domain QA datasets given the availability of other NLP resources. We also discuss possible extensions of the framework to increase the complexity of the generated datasets.
%We also discuss how the emrQA dataset can be further extented to make it more general and representative of physician QA needs.

%as an alternate to existing techniques such as crowdsourcing. 

%Our proposed frameworks ability to generate a challenging dataset has so far been established through emrQAs dataset complexity (Section  \ref{baselines} and \ref{analysis}). In Section \ref{emrqa} and \ref{frames}, we have also discussed how structured domain knowledge in logical forms can scale the expertise of annotators to new unannotated documents, allowing large-scale dataset generation in complex domains such as EMRs. To widen the scope of the frameworks application capability, we briefly outline how the process may be applied to any open-domain QA dataset generation, as an alternate to existing techniques such as crowdsourcing. 

\textbf{Open domain QA dataset generation:} Consider the popularly used SQuAD \cite{rajpurkar2016squad} reading comprehension dataset generated by crowdworkers, where the answer to every question is a segment of text from the corresponding passage in the Wikipedia  article. This dataset can easily be generated or extended using our proposed framework with existing NLP annotations on Wikipedia \cite{auer2007dbpedia, nothman2008transforming, ghaddar2017winer}.
%This can alternately be generated/or extended by using our proposed framework with existing NLP datasets on Wikipedia \cite{auer2007dbpedia, nothman2008transforming, ghaddar2017winer}. 

For instance, consider DBPedia \cite{auer2007dbpedia}, an existing dataset of entities and their relations extracted from  Wikipedia. It also has its own ontology which can serve as the semantic frames schema to define logical forms. Using these resources, our reverse engineering technique for QA dataset generation can be applied as follows. (1) Question templates can be defined for each entity type and relation in DBPedia. For example\footnote{example reference: \href{http://dbpedia.org/page/Normandy}{http://dbpedia.org/page/Normandy}}, consider the relation \textit{[place, country]} field in DBpedia. For this we can define a question template \textit{In what country is $|place|$ located?}. (2) Every such question template  can be annotated with a logical form template using existing DBPedia ontology.  (3) By considering the entity values of DBPedia fields such as \textit{[place=Normandy, dbo:country=France]}, we can automatically generate the question \textit{In what country is Normandy located?} and its corresponding logical form from the templates. The text span of \textit{country=France} from the Wikipedia passage is then used as the answer \cite{isem2013daiber}. Currently, this QA pair instance is a  part of the SQuAD dev set. Using our framework we can generate many more instances like this example from different Wikipedia passages - without crowdsourcing efforts.
%Using our framework we can generate many more instances of this example from different Wikipedia passages, without crowdsourcing efforts. 

%In this case, the answer will be the following span of text: "Robert Nesta "Bob" Marley, OM (6 February 1945 – 11 May 1981)", which would have been the span identified by crowdworker. 

\textbf{Extensions to the framework:} The complexity of the generated dataset can be further extended as follows. (1) We can use a coreferred or a lexical variant of the  original entity  in the question-logical form generation. This can allow for increased lexical variation between the question and answer line entities in the passage. (2)  It is possible to combine two or more question templates to make compositional questions with the answers to these questions similarly combined. This can also result in more multiple sentence reasoning questions. (3) We can generate questions with entities not related to the context in the passage. This can increase empty answer questions in the dataset, resulting in increased negative training examples.

%% file: conclusion.tex
\vspace{-0.05in}
\section{Conclusions and Future Work} 
\vspace{-0.04in}
%\vspace{-0.1in} 
We propose a novel framework that can generate a large-scale QA dataset using existing resources and minimal expert input. This has the potential to make a huge impact in domains like medicine, where obtaining manual QA annotations is tedious and infeasible. We apply this framework to generate a large scale EMR QA corpus (emrQA), consisting of 400,000 question-answers pairs and 1 million question-logical forms, and analyze the complexity of the dataset to show its non-trivial nature. We show that the logical forms provide a symbolic representation that is very useful for corpus generation and for model analysis. The logical forms also provide an opportunity to build interpretable systems by perhaps jointly (or latently) learning the logical form and answer for a question.  In future, this framework may be applied to also re-purpose and integrate other NLP datasets such as MIMIC and generate a more diverse and representative EMR QA corpus \cite{johnson2016mimic}.